\documentclass[lettersize,journal]{IEEEtran}

\newcommand\highlightReference[1]{%
  \expandafter\newcommand\csname highlightReference-#1\endcsname{}%
}
\let\oldbibitem\bibitem
\def\bibitem#1 #2\par{%
  \expandafter\ifx\csname highlightReference-#1\endcsname\relax
    \oldbibitem{#1}#2\par
  \else
    \oldbibitem{#1}\highlight{#2}\par
  \fi
}
\usepackage{color,soul}
\newcommand\highlight[1]{\hl{#1}}

\usepackage{amsmath,amsfonts}
\usepackage{array}
\usepackage{textcomp}
\usepackage{url}
\usepackage{verbatim}
\usepackage{graphicx}
\def\BibTeX{{\rm B\kern-.05em{\sc i\kern-.025em b}\kern-.08em
    T\kern-.1667em\lower.7ex\hbox{E}\kern-.125emX}}
\usepackage{balance}
\usepackage{color}
\usepackage{graphicx}

\usepackage{graphicx}
\usepackage{epstopdf}
\usepackage{subfigure}
\usepackage{booktabs}
\usepackage{booktabs}
\usepackage{array}
\usepackage{ragged2e} 
\usepackage{booktabs,makecell, multirow, tabularx}
\usepackage{amsmath}  
\usepackage{amssymb}  
\usepackage{romannum}
\usepackage{soul}
\usepackage{hyperref}
\usepackage{comment}

\makeatletter  
\newif\if@restonecol  
\makeatother

\usepackage[linesnumbered,ruled,vlined]{algorithm2e}

\usepackage{bm} 

\usepackage{cite}

\makeatletter

\newcommand{\Rmnum}[1]{\expandafter\@slowromancap\romannumeral #1@}
\makeatother

\usepackage{lineno}

\usepackage[table,xcdraw]{xcolor}  

\usepackage{soul, color, xcolor}
\soulregister{\cite}{7}
\soulregister{\ref}{7}
\soulregister{\eqref}{7}
\soulregister{\textit}{7}
\usepackage{lineno}

\usepackage{xpatch}

\makeatletter
\xpatchcmd{\IEEEbiography}{plus 1fil}{}{}{}
\xpatchcmd{\endIEEEbiography}{plus 1fil}{}{}{}
\xpatchcmd{\IEEEbiographynophoto}{plus 1fil}{}{}{}
\xpatchcmd{\endIEEEbiographynophoto}{plus 1fil}{}{}{}

\def\@IEEEBIOskipN{3.50\baselineskip}
\makeatother

\begin{document}
\pagenumbering{arabic}   
\setcounter{page}{1}     
 

\title{AMR-Pose: An Active LED Marker-Based Relative Pose Estimation Framework With Probabilistic Switching PnP for Cooperative AUVs}
\author{Zeyu Sha\textsuperscript{$\dagger$}, Xiaorui Wang\textsuperscript{$\dagger$}, Mingyang Yang and Feitian Zhang*

\thanks{\textsuperscript{$\dagger$} These authors contributed equally to this work.

The authors are with the Robotics and Control Laboratory, School of Advanced Manufacturing and Robotics, and the State Key Laboratory of Turbulence and Complex Systems,  Peking University, Beijing, 100871, China (email: {schahzy@stu.pku.edu.cn}, {jnswxr@stu.pku.edu.cn}, {mingyangyang@stu.pku.edu.cn}, {feitian@pku.edu.cn}).}

}

\maketitle

\begin{abstract}
Reliable relative pose estimation between autonomous underwater vehicles (AUVs) is critical for cooperative ocean exploration, sampling, and multi-robot coordination. However, achieving robust vision-based relative localization in underwater environments remains challenging due to severe optical degradation, including turbidity, illumination variations, reflections, and intermittent feature occlusions. This paper presents AMR-Pose, an active LED marker-based relative pose estimation framework for cooperative AUVs. A compact marker module consisting of one red central LED and three blue peripheral LEDs is developed and integrated onto the leader AUV to provide distinctive visual features under complex underwater conditions. Building upon the detected marker observations, a probabilistic switching Perspective-n-Point estimator (PSwPnP) is developed by combining Lie-group pose propagation on $SE(3)$, probabilistic marker association, and visibility-adaptive measurement fusion for robust six-degree-of-freedom relative pose estimation. The proposed framework dynamically adapts the estimation process according to marker visibility, maintaining geometric consistency and temporal stability during partial observations and visibility transitions. Extensive water-tank experiments with motion-capture ground truth validate that AMR-Pose achieves accurate, smooth, and robust relative pose estimation under challenging underwater conditions. Closed-loop leader-follower experiments further demonstrate its feasibility for real-time relative pose feedback in cooperative underwater robotics.
\end{abstract}

\begin{IEEEkeywords}
Autonomous underwater vehicles, underwater LED markers, relative pose tracking, probabilistic switching PnP, vision-based following.
\end{IEEEkeywords}

\section{Introduction}

\IEEEPARstart{W}{ith} the rapid development of marine robotics, autonomous underwater vehicles (AUVs) \cite{i-auv,auv-intro,auv-intro2} have been widely deployed for underwater exploration \cite{underwater-mani,underwater-mani2,mani2}, infrastructure inspection \cite{underwaterinspection,underwaterinspection3}, and environmental monitoring \cite{seabed2}. As underwater missions become increasingly complex, cooperative multi-AUV systems are increasingly adopted, making reliable relative pose estimation between vehicles a fundamental capability \cite{auv-navi,auv-intro3,mit}.

However, obtaining accurate relative pose estimates between underwater vehicles remains challenging.  Acoustic sensing has been widely adopted for underwater ranging, positioning, and target localization \cite{acoustic5,sonar}. For close-range cooperative tasks such as leader--follower navigation and formation control, continuous six-degree-of-freedom (6-DoF) relative pose estimation motivates vision-based sensing with richer geometric information. Underwater vision has been extensively investigated for perception and environment-based localization \cite{visual2-tro,visual3,visual4,visual6}, including visual SLAM  \cite{underwater-slam,underwater-slam2,slam-tro}. However, these approaches rely on environmental features and remain vulnerable to turbidity, illumination variations, reflections, and limited visual texture. More importantly, estimating the pose of a maneuvering AUV differs fundamentally from environment localization because the target itself introduces changing viewpoints, self-occlusion, and intermittent feature visibility. Therefore, reliable AUV-to-AUV relative pose estimation requires a dedicated target observation mechanism beyond conventional passive visual perception.

Active visual markers provide an effective solution for improving target observability by generating distinguishable visual features independent of environmental texture. For mobile underwater robots, however, such a sensing strategy requires not only a compact and deployable marker system but also an estimation algorithm capable of handling dynamic visibility changes caused by relative motion. Hernandez et al. \cite{Nasa} developed an LED-based underwater motion capture system using multi-camera triangulation, demonstrating the effectiveness of active illumination for accurate underwater motion measurement. 
Similar concepts have been investigated in underwater robotic applications, including marker-based docking localization \cite{Zhao}, LED-camera-based pose measurement for robot swarms \cite{Kitano}, adaptive visual positioning \cite{Li}, and LED-based vision navigation \cite{Xu}.
These studies demonstrate the potential of active markers for underwater localization. However, existing systems generally rely on fixed external cameras, docking stations, or controlled marker configurations, where marker visibility and correspondence remain relatively stable. Their applicability to mobile AUV-to-AUV relative pose estimation therefore remains limited.

Once active markers provide reliable artificial observations, the remaining challenge is to recover the relative 6-DoF pose accurately and continuously from these observations. Marker-based relative pose estimation is commonly formulated as a Perspective-n-Point (PnP) problem, where the relative pose is estimated from known three-dimensional (3D) marker geometry and corresponding image observations. EPnP \cite{epnp} provides an efficient solution for general camera pose estimation, while uncertainty-aware variants further improve robustness under noisy observations \cite{Florian,Xie,Zhan}. However, most existing PnP methods assume reliable feature correspondence and sufficient visible points, which may not hold in underwater AUV-to-AUV scenarios. Relative motion between vehicles causes marker self-occlusion, intermittent feature disappearance and reappearance, and ambiguous detections caused by underwater reflections and clutter. Consequently, robust underwater AUV-to-AUV relative pose estimation requires a co-designed sensing and estimation framework that combines reliable marker observability with visibility-adaptive pose estimation.

\begin{figure}[htbp]
    \centering
    \includegraphics[width=\linewidth, keepaspectratio]{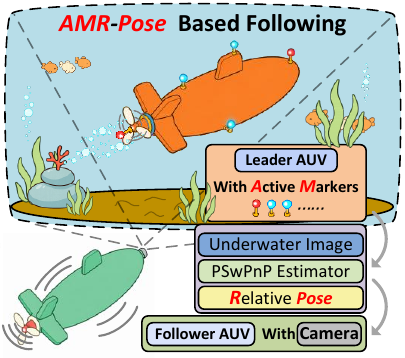}
    \caption{Overview of the proposed AMR-Pose framework. AMR-Pose integrates an active LED marker module with the PSwPnP estimator to achieve robust underwater relative pose estimation under dynamic visibility conditions.}
    \label{fig:main_idea}
\end{figure}

To address these challenges, this paper proposes AMR-Pose, an active LED marker-based relative pose estimation framework for cooperative AUVs. The proposed framework consists of two tightly coupled components: a compact active LED marker module for robust target observation and a probabilistic switching Perspective-n-Point estimator (PSwPnP) for adaptive relative pose estimation. The LED marker module (illustrated in Fig.~\ref{fig:main_idea}), consisting of one red central LED and three blue peripheral LEDs, provides distinctive sparse visual features for reliable observations. Based on these observations, PSwPnP formulates pose estimation on the Lie group $SE(3)$ \cite{Sola2018MicroLie,Barfoot2017StateEstimation} and integrates temporal pose propagation, probabilistic marker association, and visibility-adaptive measurement fusion. By dynamically adapting the estimation strategy according to marker visibility, PSwPnP maintains geometric consistency and temporal stability during partial observations and visibility transitions.

The proposed AMR-Pose framework is validated through water-tank experiments with two OpenAUV platforms and motion-capture ground truth. Comprehensive evaluations include relative pose accuracy analysis, component-removal ablation studies, dedicated $4$--$3$--$4$ visibility transition experiments, and closed-loop leader-follower experiments using real-time visual pose feedback.

The main contributions of this work are summarized as follows. First, an integrated active-marker relative pose estimation framework, AMR-Pose, is developed for cooperative AUVs. A compact red-blue LED marker module is designed and integrated onto the leader AUV to provide distinguishable visual observations under challenging underwater conditions. Second, a probabilistic switching PnP estimator (PSwPnP) is proposed by integrating Lie-group pose propagation on $SE(3)$, probabilistic marker association, and visibility-adaptive measurement fusion, enabling continuous and robust relative pose estimation under complete and partial marker observations. Furthermore, an existence-aware visibility management strategy is developed to preserve marker identity consistency and suppress unreliable measurements caused by  clutter, missed detections, and intermittent occlusions. Finally, extensive water-tank experiments with motion-capture ground truth validate AMR-Pose through pose accuracy evaluation, ablation studies, visibility transition analysis, and closed-loop leader-follower demonstrations via real-time visual feedback.

\section{Active Marker Perception System}
The perception layer of the proposed AMR-Pose framework consists of a compact active LED marker module and a lightweight image-processing pipeline that extracts reliable marker observations for the subsequent pose estimator. The active marker module enhances target observability under degraded underwater conditions, while the observation pipeline converts monocular images into sparse LED centroid measurements for robust relative pose estimation.

\subsection{Active LED Marker Module}

Reliable marker observations form the foundation of the proposed relative pose estimation framework. To provide distinctive visual features under challenging underwater conditions, a compact active LED marker module is developed and integrated onto the leader AUV. The marker configuration consists of one red central LED and three blue peripheral LEDs, enabling robust color-based identification while maintaining a compact mechanical footprint.

\begin{figure*}[t]
    \centering
    \includegraphics[width=\linewidth, keepaspectratio]{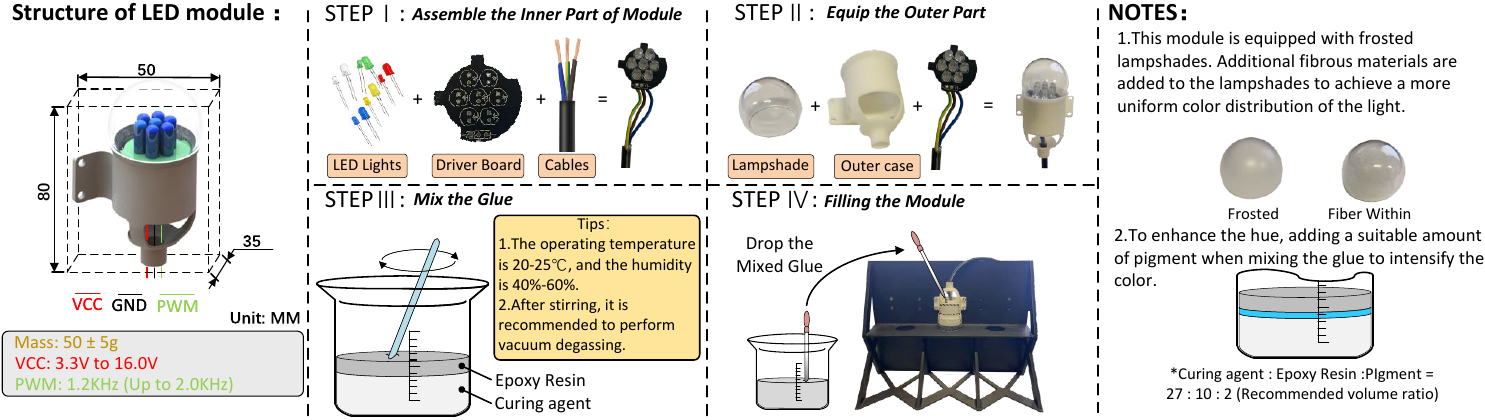}
    \caption{Design and fabrication of the proposed active LED marker module. The module integrates a pressure-resistant housing, LED emitters, a driver circuit, and waterproof encapsulation for long-term underwater deployment.}
    \label{LED}
\end{figure*}

As illustrated in Fig.~\ref{LED}, the proposed module consists of LED emitters, a driver circuit, a lampshade, and a pressure-resistant housing. The internal cavity is fully encapsulated using waterproof potting material, significantly improving pressure resistance and long-term operational reliability during underwater deployment. The modular mechanical design simplifies fabrication and assembly while facilitating straightforward maintenance and replacement.

\begin{figure}[htbp]
    \centering
    \includegraphics[width=\linewidth, keepaspectratio]{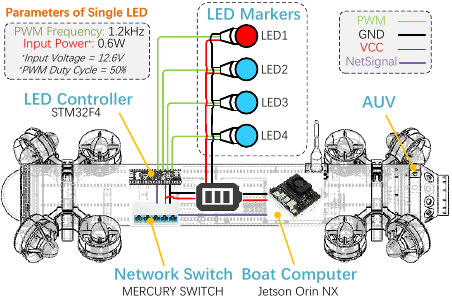}
    \caption{System integration of the active LED marker module with the AUV. The onboard computer communicates with the LED controller through an MCU to enable real-time brightness control and status management.}
    \label{AUVwithLED}
\end{figure}

The LED module supports an input voltage range of 3--16~V and provides PWM-based brightness control, enabling the illumination intensity to be adjusted according to varying water conditions and observation distances. Multiple marker modules are controlled simultaneously through an MCU-based control architecture. As illustrated in Fig.~\ref{AUVwithLED}, the MCU communicates with the onboard computer through a network interface, allowing real-time brightness regulation and device management. Standardized mounting holes and waterproof connectors further enable rapid installation and flexible deployment on different underwater robotic platforms. 
Compared with existing underwater LED beacon systems, the proposed design emphasizes compact integration, pressure resistance, adaptive illumination, and seamless compatibility with onboard robotic systems. These characteristics make the marker module well suited for long-term underwater relative pose estimation and cooperative AUV applications.

\subsection{Marker Observation Extraction}
The active LED marker module provides sparse visual observations for the subsequent PSwPnP estimator. Given an undistorted monocular image, visible LEDs are extracted by HSV-based color segmentation followed by morphological filtering and connected-component analysis to suppress underwater noise and reflections.
The central red LED is distinguished from the peripheral blue LEDs using color constraints, yielding the image coordinates of one red LED and the visible blue LEDs for subsequent marker association and relative pose estimation.

\begin{figure*}[t]
    \centering
    \includegraphics[width=\linewidth, keepaspectratio]{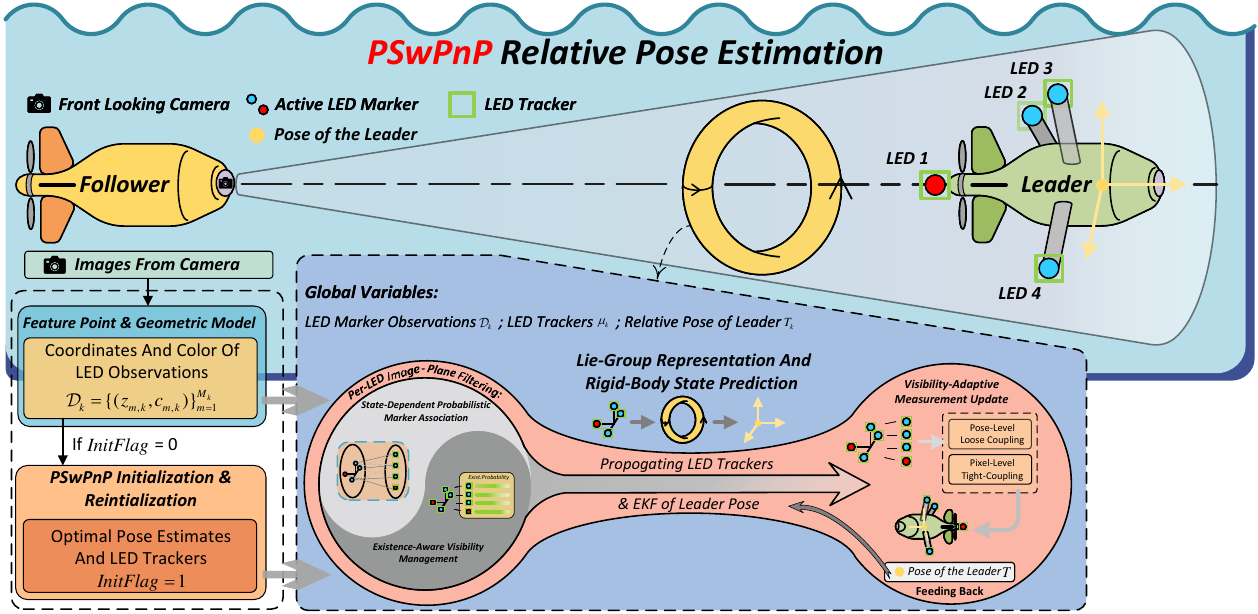}
    \caption{Overview of the proposed AMR-Pose framework. The estimator integrates Lie-group pose prediction, probabilistic marker association, existence-aware visibility management, and visibility-adaptive measurement updates to achieve robust relative pose estimation under dynamic marker visibility.}
    \label{Ele-big}
\end{figure*}

\section{PSwPnP: Probabilistic Switching Pose Estimation}
\label{sec:pswpnp}

This section presents PSwPnP, the probabilistic switching pose estimator that forms the estimation layer of the proposed AMR-Pose framework. PSwPnP integrates Lie-group motion prediction, probabilistic marker association, existence-aware visibility management, and visibility-adaptive measurement updates within a unified Bayesian estimation framework. By jointly exploiting geometric motion priors and image observations, the estimator maintains continuous 6-DoF relative pose estimates under challenging underwater conditions. According to the number of reliable marker observations, PSwPnP automatically switches between a pose-level update based on PnP and a pixel-level update based on direct image measurements, as illustrated in Fig.~\ref{Ele-big}.

\subsection{Problem Formulation}

The objective of PSwPnP is to estimate the 6-DoF relative pose of the leader AUV from observations of the active LED marker array. The marker array consists of four calibrated LEDs, and the target body frame is defined with its origin at the array center.

Let $P_i\in\mathbb{R}^3$ denote the known position of LED $i$ in the target body frame, and let $T_k\in SE(3)$ denote the relative pose of the leader with respect to the follower at time step $k$. The follower carries a calibrated monocular camera with extrinsic transformation $T_{cf}\in SE(3)$ and intrinsic matrix $K$. The coordinate of LED $i$ in the camera frame is
\begin{equation}
\begin{bmatrix}
p_{i,k}^{c}\\
1
\end{bmatrix}
=
T_{cf}T_k
\begin{bmatrix}
P_i\\
1
\end{bmatrix}.
\end{equation}
For $p^c=[X~Y~Z]^\top$, the image projection is
\begin{equation}
\pi(p^c)=
\begin{bmatrix}
f_x X/Z + c_x \\
f_y Y/Z + c_y
\end{bmatrix}.
\end{equation}
The perception module provides a set of LED observations, i.e.,
\begin{equation}
    \mathcal{D}_k=\{(z_{m,k},c_{m,k})\}_{m=1}^{M_k},
\end{equation}
where $z_{m,k}\in\mathbb{R}^2$ is the image centroid of detection $m$ and $c_{m,k}$ is the corresponding color label. The objective of PSwPnP is to estimate $T_k$ from $\mathcal{D}_k$ over time.

\subsection{Initialization}

Initialization identifies the correct LED correspondences and obtains the initial relative pose. Because underwater reflections and clutter may produce multiple red and blue candidates, PSwPnP formulates initialization as a color-consistent hypothesis search. Let $\mathcal{D}_r$ and $\mathcal{D}_b$ denote the red and blue candidate sets, respectively. Each hypothesis $c\in\mathcal{C}$ consists of one red candidate and an ordered triplet of distinct blue candidates,
\begin{equation}
c=(q_r,q_{b_1},q_{b_2},q_{b_3}),
\qquad
q_r\in\mathcal{D}_r,\quad
q_{b_\ell}\in\mathcal{D}_b,
\end{equation}
where the ordered blue candidates correspond to the three peripheral LEDs and $q_{b_\alpha}\ne q_{b_\beta}$ for $\alpha\ne\beta$.

For each hypothesis, an initial pose is computed using EPnP and refined by Levenberg--Marquardt. Denoting the refined pose by $(R_c,t_c)$ and the image observation assigned to LED $i$ by $q_i(c)$, the reprojection error is
\begin{equation}
r_i(c)=
\left\|
q_i(c)-\pi\left(R_c P_i+t_c\right)
\right\|.
\end{equation}
A hypothesis score is defined as
\begin{equation}
\operatorname{score}(c)=
\exp\left(
-\sum_{i=1}^{4}
\frac{r_i(c)^2}{2\sigma_{\mathrm{pix}}^2}
\right).
\end{equation}
The optimal hypothesis is selected as
\begin{equation}
c^\ast=\arg\max_{c\in\mathcal{C}}\operatorname{score}(c).
\end{equation}
To suppress false initialization under clutter, the normalized confidence is computed as
\begin{equation}
P(c^\ast\mid\mathcal{D})=
\frac{\operatorname{score}(c^\ast)}
{\sum_{c\in\mathcal{C}}\operatorname{score}(c)+C\lambda_{\mathrm{clutter}}},
\end{equation}
where $C=|\mathcal{C}|$ is the number of hypotheses and $\lambda_{\mathrm{clutter}}$ is the clutter coefficient. The solution is accepted only if $P(c^\ast\mid\mathcal{D})$ exceeds a predefined threshold. The accepted hypothesis initializes both the pose estimator and the four LED trackers, providing a consistent initialization for subsequent estimation.

Reinitialization is triggered when marker association fails or the reliable marker set collapses. If successful, the pose estimate and marker identities are refreshed using the same hypothesis selection procedure. Otherwise, the estimator preserves the previous assignment and restores the predicted filter states, preventing transient false detections from disrupting a stable solution.

\subsection{Lie-Group Motion Prediction}

The relative leader pose is represented on the Lie group $SE(3)$ instead of using a Euclidean pose vector \cite{Sola2018MicroLie,Barfoot2017StateEstimation}. A pose is written as
\begin{equation}
SE(3)=\left\{
T=
\begin{bmatrix}
R & t\\
0 & 1
\end{bmatrix}
~\middle|~
R\in SO(3),~t\in\mathbb{R}^3
\right\},
\end{equation}
where $R$ and $t$ denote rotation and translation, respectively. The associated Lie algebra $\mathfrak{se}(3)$ is parameterized by
\begin{equation}
\xi=
\begin{bmatrix}
\rho\\
\phi
\end{bmatrix}\in\mathbb{R}^6,
\qquad
\xi^\wedge=
\begin{bmatrix}
\phi^\wedge & \rho\\
0 & 0
\end{bmatrix},
\end{equation}
where $\phi^\wedge$ is the skew-symmetric matrix of $\phi$, and $(\xi^\wedge)^\vee=\xi$. The exponential and logarithm maps define the local correspondence between $SE(3)$ and $\mathfrak{se}(3)$,
\begin{equation}
T=\operatorname{Exp}(\xi^\wedge),
\qquad
\xi=(\operatorname{Log}(T))^\vee .
\end{equation}
The pose is parameterized by exponential coordinates, while the state perturbation and covariance are represented additively in the corresponding six-dimensional coordinate space. The left Jacobian is used to relate additive coordinate perturbations to local group perturbations during geometric linearization.

For $T\in SE(3)$, the adjoint matrix is \cite{Sola2018MicroLie}
\begin{equation}
\operatorname{Ad}_T=
\begin{bmatrix}
R & [t]_\times R\\
0 & R
\end{bmatrix}.
\end{equation}
The left Jacobian $J_\ell(\cdot)$ is defined by the first-order relation \cite{Sola2018MicroLie}
\begin{equation}
\operatorname{Exp}\!\left((\eta+\delta)^\wedge\right)
\approx
\operatorname{Exp}\!\left((J_\ell(\eta)\delta)^\wedge\right)
\operatorname{Exp}(\eta^\wedge),
\end{equation}
where $\eta$ is the nominal tangent vector and $\delta$ is a small perturbation.

The rigid-body state is defined as
\begin{equation}
x_k=
\begin{bmatrix}
\xi_k^\top & v_k^\top & a_k^\top
\end{bmatrix}^\top,
\end{equation}
where $\xi_k\in\mathbb{R}^6$ parameterizes the pose, and $v_k,a_k\in\mathbb{R}^6$ denote twist and twist acceleration. The spatial twist satisfies the kinematic relation
$\dot{T}(t)=v^\wedge T(t)$. With sampling interval $\Delta t$, the discrete-time prediction model is
\begin{equation}
\label{eq:pswpnp_pred_pose}
\begin{gathered}
\Delta_k=v_k \Delta t+\frac{1}{2}a_k\Delta t^2, \\
\hat{T}_{k+1|k}
=
\operatorname{Exp}\!\left(\Delta_k^\wedge\right)
\hat{T}_{k|k},\\
\hat{\xi}_{k+1|k}
=
\operatorname{Log}\!\left(
\operatorname{Exp}\!\left(\Delta_k^\wedge\right)
\operatorname{Exp}\!\left(\hat{\xi}_{k|k}^{\wedge}\right)
\right)^\vee .
\end{gathered}
\end{equation}
The velocity and acceleration evolve as
\begin{equation}
\label{eq:pswpnp_pred_va}
\begin{gathered}
v_{k+1|k}=v_k+a_k\Delta t+w_{v,k},\\
a_{k+1|k}=a_k+w_{a,k},
\end{gathered}
\end{equation}
where $w_{v,k}\sim\mathcal{N}(0,Q_v)$ and $w_{a,k}\sim\mathcal{N}(0,Q_a)$.

We next derive the prediction Jacobian for covariance propagation. Define
\begin{equation}
\label{eq:pswpnp_group_def}
\begin{gathered}
T_\Delta=\operatorname{Exp}\!\left(\Delta_k^\wedge\right),\qquad
T_\xi=\operatorname{Exp}\!\left(\xi_k^\wedge\right),\\
T=T_\Delta T_\xi=\operatorname{Exp}\!\left(\psi^\wedge\right),
\qquad
\psi=(\operatorname{Log}T)^\vee .
\end{gathered}
\end{equation}
For small perturbations $\delta\xi$ and $\delta\Delta$, the left-Jacobian relation, the conjugate identity
\[
T_\Delta\operatorname{Exp}(u^\wedge)
=
\operatorname{Exp}\!\left((\operatorname{Ad}_{T_\Delta}u)^\wedge\right)T_\Delta,
\]
and a first-order BCH approximation give
\begin{equation}
\label{eq:pswpnp_Tprime_bch}
T^\prime
\approx
\operatorname{Exp}\!\left(
\left[
J_\ell(\Delta_k)\delta\Delta
+
A_\Delta J_\ell(\xi_k)\delta\xi
\right]^\wedge
\right)T,
\end{equation}
where $A_\Delta=\operatorname{Ad}_{T_\Delta}$. Comparing \eqref{eq:pswpnp_Tprime_bch} with the left-Jacobian perturbation of $\operatorname{Exp}(\psi^\wedge)$ yields
\begin{equation}
\label{eq:pswpnp_dpsi}
\delta\psi
=
J_\ell(\psi)^{-1}
\left(
J_\ell(\Delta_k)\delta\Delta
+
A_\Delta J_\ell(\xi_k)\delta\xi
\right).
\end{equation}
Therefore, the pose-transition partial derivatives are
\begin{equation}
\label{eq:pswpnp_partials}
\begin{gathered}
\frac{\partial\psi}{\partial\xi_k}
=
J_\ell(\psi)^{-1}A_\Delta J_\ell(\xi_k),\\
\frac{\partial\psi}{\partial\Delta_k}
=
J_\ell(\psi)^{-1}J_\ell(\Delta_k).
\end{gathered}
\end{equation}
Since $\Delta_k=v_k\Delta t+\frac{1}{2}a_k\Delta t^2$, the chain rule gives
\begin{equation}
\label{eq:pswpnp_dpsi_va}
\begin{gathered}
\frac{\partial\psi}{\partial v_k}
=
J_\ell(\psi)^{-1}J_\ell(\Delta_k)\Delta t,\\
\frac{\partial\psi}{\partial a_k}
=
J_\ell(\psi)^{-1}J_\ell(\Delta_k)\frac{1}{2}\Delta t^2.
\end{gathered}
\end{equation}

Define the error vector $\delta x=[\delta \xi^\top~\delta v^\top~\delta a^\top]^\top \in \mathbb{R}^{18}$. The state transition Jacobian is
\begin{equation}
\label{eq:pswpnp_F}
F_k=
\begin{bmatrix}
F_{\xi \xi,k} & F_{\xi v,k} & F_{\xi a,k} \\
0 & I_6 & \Delta t I_6 \\
0 & 0 & I_6
\end{bmatrix},
\end{equation}
where
\begin{equation}
\label{eq:pswpnp_Fblocks}
\begin{gathered}
F_{\xi \xi,k}=J_\ell(\psi)^{-1} A_\Delta J_\ell(\xi_k),\\
F_{\xi v,k}=J_\ell(\psi)^{-1} J_\ell(\Delta_k)\Delta t,\\
F_{\xi a,k}=J_\ell(\psi)^{-1} J_\ell(\Delta_k)\frac{1}{2}\Delta t^2.
\end{gathered}
\end{equation}
The process noise transfer matrix and covariance prediction follow the standard Lie-group uncertainty propagation form \cite{Barfoot2014AssociatingUncertainty,Barfoot2017StateEstimation},
\begin{equation}
\label{eq:pswpnp_G}
G_k=
\begin{bmatrix}
F_{\xi v,k} & F_{\xi a,k} \\
I_6 & 0 \\
0 & I_6
\end{bmatrix},
\qquad
Q_k=\operatorname{blkdiag}(Q_v,Q_a),
\end{equation}
\begin{equation}
\label{eq:pswpnp_Ppred}
P_{k+1|k}=F_kP_{k|k}F_k^\top+G_kQ_kG_k^\top .
\end{equation}
For subsequent reprojection and association, only the predicted pose $\hat{T}_{k+1|k}$ and the corresponding pose covariance block $P_{\xi\xi,k+1|k}$ are required.

\subsection{Probabilistic Marker Association}
\subsubsection{Image-Plane Prediction}

After initialization, each physical LED is associated with an independent image-plane Kalman filter that predicts its pixel location in subsequent frames. These filters provide short-term motion priors for probabilistic marker association and support temporal continuity during brief missed detections or occlusions. For tracker $j$, let $\mu_{j,k}^{kf+}$ and $\Sigma_{j,k}^{kf+}$ denote the posterior mean and covariance of its pixel state in frame $k$. The prediction step follows
\begin{equation}
\mu_{j,k+1}^{kf-}=A_j\,\mu_{j,k}^{kf+},
\qquad
\Sigma_{j,k+1}^{kf-}=A_j\,\Sigma_{j,k}^{kf+}A_j^\top+Q_{j}^{kf},
\end{equation}
where $A_j$ is the image-plane state transition matrix and $Q_j^{kf}$ is the process covariance. If tracker $j$ is active, its image-plane prediction $\bigl(\mu_{j,k}^{kf-},\Sigma_{j,k}^{kf-}\bigr)$ is used for association. Otherwise, association falls back to the rigid-body reprojection prior induced by $\hat{T}_{k|k-1}$ and $P_{\xi\xi,k|k-1}$. 

\subsubsection{Tracker-State-Dependent Association}

To maintain marker identity consistency under intermittent visibility, PSwPnP performs  probabilistic association between the current detections and the four LED trackers. The key idea is to adapt the association prior according to the tracker state. 
For LED $j$ with target-frame coordinate $P_j$, the reprojection prior induced by the rigid-body prediction is
\begin{equation}
\begin{bmatrix}
\bar{p}_{j,k}^{c}\\
1
\end{bmatrix}
=
T_{cf}\hat{T}_{k|k-1}
\begin{bmatrix}
P_j\\
1
\end{bmatrix},
\mu_{j,k}^{r}=\pi\!\left(\bar{p}_{j,k}^{c}\right),
\end{equation}
with covariance
\begin{equation}
\Sigma_{j,k}^{r}=
J_{\text{pose}\rightarrow\text{pix},j,k}\,
P_{\xi\xi,k|k-1}\,
J_{\text{pose}\rightarrow\text{pix},j,k}^{\top}
+
R_{\text{proj}},
\end{equation}
where $J_{\text{pose}\rightarrow\text{pix},j,k}$ is the reprojection Jacobian with respect to the additive exponential-coordinate state $\xi$, evaluated at $\hat{\xi}_{k|k-1}$, and $R_{\text{proj}}$ is the projection noise covariance.

The association prior $(\mu_{j,k},\Sigma_{j,k})$ is selected from either the image-plane prediction or the reprojection prediction according to the tracker state.
For detection $i$ with pixel coordinate $p_{i,k}$, the association likelihood is
\begin{equation}
L_{ij,k}=\mathcal{N}\!\left(p_{i,k};\mu_{j,k},\Sigma_{j,k}\right).
\end{equation}
For numerical stability, the log-likelihood is used, i.e.,
\begin{align}
\log L_{ij,k}
& =
-\frac{1}{2}\left(p_{i,k}-\mu_{j,k}\right)^{\top}
\Sigma_{j,k}^{-1}
\left(p_{i,k}-\mu_{j,k}\right) \nonumber \\ 
& -\frac{1}{2}\log |\Sigma_{j,k}|-\log(2\pi),
\end{align}
while unmatched detections are modeled using a clutter likelihood $L_{0j,k}=\lambda_{\text{clutter}}$. The association probability is normalized using log-sum-exp, i.e.,
\begin{equation}
\log Z_{j,k}
=
\ell_{\max,j,k}
+
\log\!\left(
\sum_{i=0}^{M_k} \exp\!\left(\ell_{ij,k}-\ell_{\max,j,k}\right)
\right),
\end{equation}
\begin{equation}
P_{ij,k}=\exp\!\left(\ell_{ij,k}-\log Z_{j,k}\right),
\qquad
\sum_{i=0}^{M_k}P_{ij,k}=1.
\end{equation}
Here, $\ell_{ij,k}=\log L_{ij,k}$ and $\ell_{\max,j,k}=\max_{0\le i\le M_k}\ell_{ij,k}$. Finally, the optimal one-to-one mapping between  detections and trackers is obtained using the Hungarian algorithm \cite{Kuhn1955Hungarian}.

\subsection{Existence-Aware Visibility Management}

To maintain stable tracking under intermittent detections, each LED tracker is assigned an existence probability. For tracker $j$, the prior is at frame $k$ is
\begin{equation}
E_{j,k}^{-}=p_{\text{survive}} \, E_{j,k-1}^{\mathrm{ctrl}},
\end{equation}
and the association evidence is
\begin{equation}
\Gamma_{j,k}=\sum_{i=1}^{M_k} L_{ij,k}.
\end{equation}
If tracker $j$ is associated a detection, the posterior existence probability is
\begin{equation}
E_{j,k}^{+}=
\frac{E_{j,k}^{-} p_D \Gamma_{j,k}}
{\left(1-E_{j,k}^{-}\right)\lambda + E_{j,k}^{-} p_D \Gamma_{j,k}}.
\end{equation}
Unmatched trackers retains the prior $E_{j,k}^{+}=E_{j,k}^{-}$.

For temporal stability, the existence probability is smoothed in the logit space, i.e.,
\begin{equation}
\ell_{j,k}^{-}=\operatorname{logit}\!\left(E_{j,k-1}^{\mathrm{ctrl}}\right),
\qquad
\ell_{j,k}^{+}=\operatorname{logit}\!\left(E_{j,k}^{+}\right),
\end{equation}
\begin{equation}
\ell_{j,k}^{\mathrm{ctrl}}=(1-\beta)\ell_{j,k}^{-}+\beta \ell_{j,k}^{+},
\qquad
E_{j,k}^{\mathrm{ctrl}}=\sigma\!\left(\ell_{j,k}^{\mathrm{ctrl}}\right).
\end{equation}

Trackers with $E_{j,k}^{\mathrm{ctrl}}>E_{\text{use}}$ are regarded as reliable and are allowed to participate in pose updates. Trackers with $E_{j,k}^{\mathrm{ctrl}}<E_{\text{delete}}$ are removed, together with their image-plane filters. Lost trackers are rebuilt when $E_{j,k}^{\mathrm{ctrl}}>E_{\text{confirm}}$.

Let $\mathcal{A}_k$ denote the reliable LED set selected by the existence module, and let $m_k=|\mathcal{A}_k|$ denote its cardinality. When $m_k=4$, PSwPnP performs a pose-level loose-coupling update. When $1\le m_k<4$, PSwPnP performs a pixel-level tight-coupling update. When $m_k=0$, only the prediction is retained.

The image-plane Kalman filter uses an association-confidence-adaptive measurement covariance for tracker $j$ assigned to detection $i^\star$, given by
\begin{equation}
R_{j,k}^{\mathrm{eff}}=
\frac{R_{\mathrm{det}}}{\max\left(P_{i^\star j,k}, \epsilon\right)}.
\end{equation}
Assuming that the pixel measurement directly observes the centroid state, the corresponding Kalman update is
\begin{equation}
K_{j,k}=
\Sigma_{j,k}^{kf-}
\left(\Sigma_{j,k}^{kf-}+R_{j,k}^{\mathrm{eff}}\right)^{-1},
\end{equation}
\begin{equation}
\mu_{j,k}^{kf+}=
\mu_{j,k}^{kf-}
+
K_{j,k}\left(z_{j,k}-\mu_{j,k}^{kf-}\right),
\end{equation}
\begin{equation}
\Sigma_{j,k}^{kf+}=
\left(I-K_{j,k}\right)\Sigma_{j,k}^{kf-}\left(I-K_{j,k}\right)^{\top}
+
K_{j,k}R_{j,k}^{\mathrm{eff}}K_{j,k}^{\top},
\end{equation}
where $z_{j,k}$ is the pixel measurement assigned to tracker $j$.

\subsection{Visibility-Adaptive Measurement Update}

\subsubsection{Pose-Level Loose-Coupling Update}

When all four LEDs are reliably observed ($m_k=4$), PSwPnP performs a pose-level update. An initial pose is obtained by EPnP and refined through Levenberg--Marquardt optimization. Define the stacked pixel residual
\begin{equation}
r_{u,k}(\xi)=
z_{\mathrm{pix},k}^{\mathrm{obs}}
-
\Pi(\xi)
\in \mathbb{R}^{8},
\end{equation}
where $z_{\mathrm{pix},k}^{\mathrm{obs}}$ stacks the four LED measurements and $\Pi(\xi)$ stacks their projections under $T_{cf}\operatorname{Exp}(\xi^\wedge)$. The refined estimate is
\begin{equation}
\hat{\xi}_{k}^{\,\mathrm{pnp}}
=
\arg\min_{\xi}
r_{u,k}(\xi)^{\top}
\Sigma_{u,k}^{-1}
r_{u,k}(\xi),
\end{equation}
where $\Sigma_{u,k}$ is the pixel measurement covariance. Since $\xi$ is the optimization variable, its covariance is approximated by
\begin{equation}
\Sigma_{\xi,k}
\approx
\left(
J_{u,k}^{\top}\Sigma_{u,k}^{-1}J_{u,k}
\right)^{-1},
\qquad
J_{u,k}=
\frac{\partial r_{u,k}}{\partial \xi}.
\label{eq:Sigxi}
\end{equation}

The pose observation is modeled as
\begin{equation}
z_{\xi,k}
=
\hat{\xi}_{k}^{\,\mathrm{pnp}}
=
\xi_k+n_{\xi,k},
\qquad
n_{\xi,k}\sim\mathcal{N}(0,\Sigma_{\xi,k}),
\end{equation}
with
\begin{equation}
H_{\mathrm{pose},k}
=
\begin{bmatrix}
I_6 & 0_{6\times6} & 0_{6\times6}
\end{bmatrix}.
\end{equation}
Let $R_{\xi,k}=\Sigma_{\xi,k}$. The innovation covariance, Kalman gain, and correction are

\begin{align}
S_{\xi,k}
&=
H_{\mathrm{pose},k}P_{k|k-1}H_{\mathrm{pose},k}^{\top}
+
R_{\xi,k},\\
K_{\xi,k}
&=
P_{k|k-1}H_{\mathrm{pose},k}^{\top}S_{\xi,k}^{-1},\\
r_{\xi,k}
=
z_{\xi,k}&-H_{\mathrm{pose},k}\hat{x}_{k|k-1},
\qquad
\delta x_k=K_{\xi,k}r_{\xi,k}.
\end{align}

The state is updated additively in the exponential-coordinate space,
\begin{equation}
\hat{x}_{k|k}
=
\hat{x}_{k|k-1}
+
\delta x_k,
\qquad
\hat{T}_{k|k}
=
\operatorname{Exp}\!\left(
\hat{\xi}_{k|k}^{\wedge}
\right),
\end{equation}
where $\hat{\xi}_{k|k}=[\hat{x}_{k|k}]_{1:6}$. The covariance follows the standard EKF update.

\subsubsection{Pixel-Level Tight-Coupling Update}

When only a subset of LEDs is reliably observed ($1\le m_k<4$), PSwPnP performs a pixel-level update. Let $h_{\mathrm{pix}}(x_k)$ denote the stacked projections of all LEDs in the reliable set $\mathcal A_k$. The observation Jacobian is
\begin{equation}
H_{\mathrm{pix},k}
=
\begin{bmatrix}
H_{\xi,k} & 0 & 0
\end{bmatrix},
\end{equation}
where
\begin{equation}
\begin{aligned}
H_{\xi,k}
&=
\begin{bmatrix}
H_{i_1\xi,k}\\
\vdots\\
H_{i_{m_k}\xi,k}
\end{bmatrix},
\qquad
H_{i\ell,k}
=
\frac{\partial \pi}{\partial p_c}
\frac{\partial p_c}{\partial\delta\xi_{\ell}},\\
H_{i\xi,k}
&=
H_{i\ell,k}
J_{\ell}\!\left(\hat{\xi}_{k|k-1}\right),
\qquad i\in\mathcal A_k.
\end{aligned}
\end{equation}

Here, $\delta\xi_{\ell}$ denotes the local left perturbation, and $J_{\ell}(\hat{\xi}_{k|k-1})$ maps the additive perturbation of $\xi$ to the corresponding local pose perturbation.

The pixel residual and innovation covariance are
\begin{equation}
r_{\mathrm{pix},k}
=
z_{\mathrm{pix},k}^{\mathrm{obs}}
-
h_{\mathrm{pix}}(\hat{x}_{k|k-1}),
\end{equation}
\begin{equation}
S_{\mathrm{pix},k}
=
H_{\mathrm{pix},k}P_{k|k-1}H_{\mathrm{pix},k}^{\top}
+
R_{\mathrm{pix},k}.
\end{equation}
The Kalman gain and correction are
\begin{equation}
K_{\mathrm{pix},k}
=
P_{k|k-1}H_{\mathrm{pix},k}^{\top}S_{\mathrm{pix},k}^{-1},
\qquad
\delta x_k
=
K_{\mathrm{pix},k}r_{\mathrm{pix},k}.
\end{equation}
The corrected state and pose are
\begin{equation}
\hat{x}_{k|k}
=
\hat{x}_{k|k-1}
+
\delta x_k,
\qquad
\hat{T}_{k|k}
=
\operatorname{Exp}\!\left(
\hat{\xi}_{k|k}^{\wedge}
\right),
\end{equation}
where $\hat{\xi}_{k|k}=[\hat{x}_{k|k}]_{1:6}$. The covariance follows the standard EKF update. If $m_k=0$, only the prediction is retained.

Algorithm~\ref{alg:pswpnp} summarizes the online pose estimation procedure of the proposed PSwPnP estimator.

\begin{algorithm}[t]
\caption{PSwPnP for Underwater Relative Pose Estimation}
\label{alg:pswpnp}
\SetKwInOut{KwIn}{Input}
\SetKwInOut{KwOut}{Output}
\KwIn{Image stream, camera model $(K,T_{cf})$, LED geometry $\{P_i\}_{i=1}^{4}$}
\KwOut{Estimated target pose $\hat{T}_{k|k}$}

Obtain detections $\mathcal{D}_1$ from LED segmentation\;
Initialize pose and marker identities using color-constrained EPnP\;
Initialize the rigid-body filter $(\hat{T}_{0|0}, P_{0|0})$ and the four image-plane trackers\;

\ForEach{frame $k=1,2,\dots$}{
    Obtain detections $\mathcal{D}_k$ from LED segmentation\;
    Predict the rigid-body state $(\hat{x}_{k|k-1},P_{k|k-1})$ using \eqref{eq:pswpnp_pred_pose}--\eqref{eq:pswpnp_Ppred}, and reconstruct $\hat{T}_{k|k-1}=\operatorname{Exp}(\hat{\xi}_{k|k-1}^{\wedge})$\;
    Predict the image-plane states of the four LED trackers to obtain
    $(\mu_{j,k}^{kf-},\Sigma_{j,k}^{kf-})$, $j=1,\dots,4$\;
    Compute the association likelihoods $L_{ij,k}$ and obtain a one-to-one assignment via Hungarian matching\;
    Update the existence probabilities $E_{j,k}^{\mathrm{ctrl}}$ and determine $\mathcal{A}_k$\;
    Update the image-plane trackers using $R_{j,k}^{\mathrm{eff}}$\;
    Compute $m_k=|\mathcal{A}_k|$\;
    
    \uIf{$m_k=4$}{
        Solve pose by EPnP and LM, compute $\Sigma_{\xi,k}$ by \eqref{eq:Sigxi}, and perform the pose-level update\;
    }
    \uElseIf{$1\le m_k<4$}{
        Construct $H_{\mathrm{pix},k}$ and perform the pixel-level tight-coupling update\;
    }
    \Else{
        Retain the prediction only, namely $\hat{T}_{k|k}=\hat{T}_{k|k-1}$\;
    }
    
    Optionally trigger reinitialization\;
    \If{reinitialization fails}{
        Preserve the previous assignment and restore the predicted filter states\;
    }
    Output $\hat{T}_{k|k}$\;
}
\end{algorithm}


\section{Experimental Validation}

The proposed AMR-Pose framework is experimentally validated using two OpenAUV\cite{OpenAUV} platforms in a water tank. The evaluation covers overall pose estimation accuracy, component ablation studies, comparison with frame-wise PnP baselines, visibility-transition experiments, and leader-follower demonstrations.

\subsection{Experimental Setup}

\begin{figure}[htbp]
    \centering
    \includegraphics[width=\linewidth, keepaspectratio]{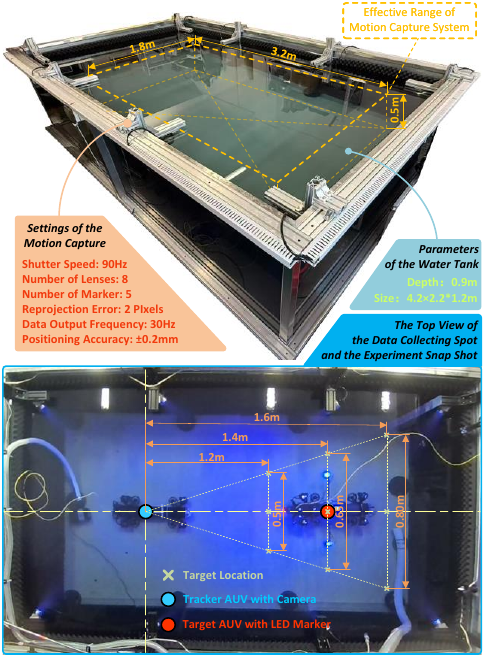}
    \caption{Experimental setup, including the water tank, underwater motion capture system, nine relative test locations, and representative experimental snapshot.}
    \label{tank}
\end{figure}

Water-tank experiments were conducted using two OpenAUV platforms, as illustrated in Fig.~\ref{tank}. The leader AUV was equipped with the proposed LED marker array, while the follower AUV carried a monocular camera. An underwater motion capture system provided ground-truth relative poses within a measurement volume of 3.2\,m $\times$ 1.6\,m $\times$ 0.5\,m.

To evaluate different viewing distances and lateral offsets, the leader was positioned at nine cone-shaped relative $(x,y)$ locations in front of the follower, namely $(1.6,-0.4)$\,m, $(1.6,0)$\,m, $(1.6,0.4)$\,m, $(1.4,-0.325)$\,m, $(1.4,0)$\,m, $(1.4,0.325)$\,m, $(1.2,-0.25)$\,m, $(1.2,0)$\,m, and $(1.2,0.25)$\,m. At each location, the leader performed a predefined 50\,s yaw maneuver consisting of hovering, $30^\circ$ left and right rotations, intermdediate returns to nominal heading. Each experiment was repeated five times, resulting in 45 trials.

Unless otherwise specified, all methods used identical camera calibration, LED geometry, image segmentation. The parameters of PSwPnP were fixed for all reported experiments. 

\subsection{Compared Methods}
The proposed PSwPnP is compared with three ablation variants and two frame-wise PnP baselines.
The ablation variants are: \emph{w/o EKF}, which removes the Lie-group rigid-body state estimator; \emph{w/o Assoc.}, which replaces probabilistic marker association with deterministic nearest-neighbor  matching; and \emph{w/o Exist.}, which replaces probabilistic existence management with hard thresholding.

Two frame-wise pose estimation methods are further considered. \emph{EPnP-LM} performs deterministic LED matching followed by EPnP and Levenberg--Marquardt refinement. \emph{GMLPnP} replaces EPnP with the generalized maximum-likelihood PnP solver \cite{Zhan}, which explicitly models anisotropic image measurement uncertainty. 
When four-marker correspondence cannot be established, the previous pose estimate is retained.

\subsection{Evaluation Metrics}

Performance is evaluated from four aspects: pose accuracy, temporal smoothness, reprojection consistency, and markder identity consistency. Pose accuracy is quantified using axis-wise root-mean-square-errors (RMSEs) together with the overall translation and rotation RMSEs, denoted by $e_t$ and $e_r$, respectively. Temporal smoothness is evaluated by the mean and 95th-percentile frame-to-frame pose increments. Reprojection consistency is measured by the pixel reprojection error, and marker identity consistency is quantified using the per-marker ID agreement rate.

\begin{figure*}[htbp]
    \centering
    \includegraphics[width=\linewidth, keepaspectratio]{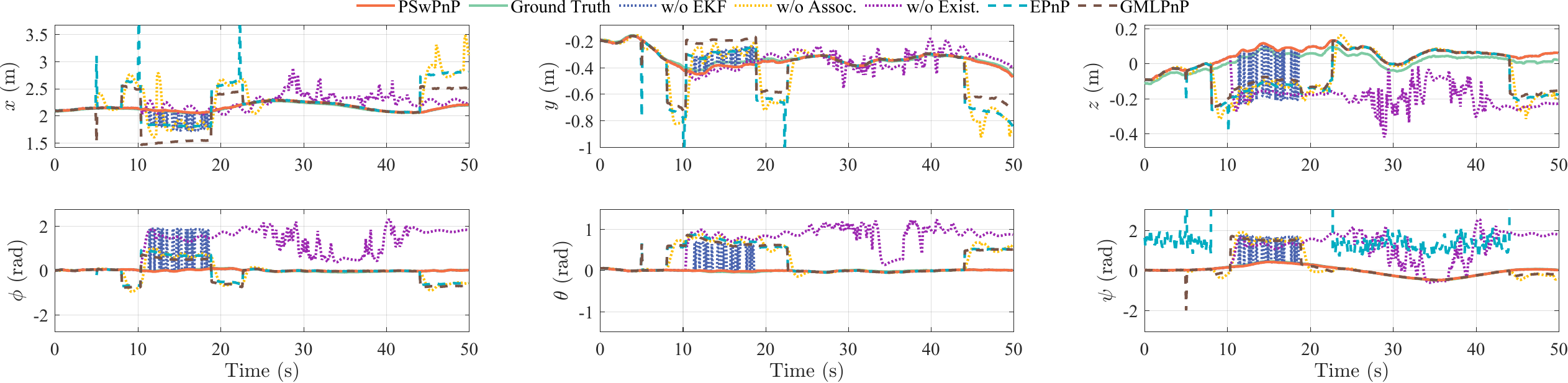}
    \caption{Representative 6-DoF relative pose estimation results at the $(1.6,-0.4)$\,m test location. The proposed PSwPnP and its ablation variants are compared with motion-capture ground truth over the complete maneuver.}
    \label{part1-1}
\end{figure*}

\begin{table*}[htbp]
\centering
\caption{Overall quantitative comparison with PnP baselines and ablation variants.}
\label{tab:overall_metrics}

\setlength{\tabcolsep}{3.0pt}
\renewcommand{\arraystretch}{1.08}
\resizebox{\textwidth}{!}{
\begin{tabular}{lccccccccccccccccccc}
\toprule
\textbf{Method} 
& \multicolumn{4}{c}{\textbf{Translation RMSE}} 
& \multicolumn{4}{c}{\textbf{Rotation RMSE}} 
& \multicolumn{4}{c}{\textbf{Smoothness}} 
& \multicolumn{2}{c}{\textbf{Reprojection}} 
& \multicolumn{5}{c}{\textbf{ID Agreement}} \\
\cmidrule(lr){2-5}
\cmidrule(lr){6-9}
\cmidrule(lr){10-13}
\cmidrule(lr){14-15}
\cmidrule(lr){16-20}
& $e_x$ & $e_y$ & $e_z$ & $e_t$
& $e_{\phi}$ & $e_{\theta}$ & $e_{\psi}$ & $e_r$
& $\bar{\Delta p}$ & $Q_{95}(\Delta p)$
& $\bar{\Delta \Theta}$ & $Q_{95}(\Delta \Theta)$
& $\bar{e}_{\mathrm{rep}}$ & $Q_{95}(e_{\mathrm{rep}})$
& $A_1$ & $A_2$ & $A_3$ & $A_4$ & $\bar{A}$ \\
\midrule

\textbf{PSwPnP}      
& \textbf{0.022} & \textbf{0.015} & \textbf{0.018} & \textbf{0.032}
& \textbf{0.022} & \textbf{0.015} & \textbf{0.017} & \textbf{0.032}
& \textbf{0.004} & \textbf{0.007} & \textbf{0.007} & \textbf{0.012}
& \textbf{1.995} & \textbf{3.191}
& \textbf{1.000} & \textbf{0.997} & \textbf{1.000} & \textbf{0.995} & \textbf{0.998} \\

\textbf{w/o EKF}    
& 0.098 & 0.031 & 0.090 & 0.137
& 0.589 & 0.192 & 0.439 & 0.587
& 0.055 & 0.460 & 0.223 & 1.932
& 1.243 & 2.233
& 1.000 & 0.996 & 0.997 & 0.995 & 0.997 \\

\textbf{w/o Assoc.}  
& 0.399 & 0.233 & 0.187 & 0.499
& 0.578 & 0.491 & 0.811 & 0.887
& 0.039 & 0.126 & 0.037 & 0.157
& 44.974 & 126.917
& 1.000 & 0.851 & 0.539 & 0.853 & 0.811 \\

\textbf{w/o Exist.}  
& 0.249 & 0.064 & 0.188 & 0.319
& 1.307 & 0.583 & 1.175 & 1.143
& 0.017 & 0.069 & 0.036 & 0.180
& 38.812 & 94.049
& 1.000 & 0.593 & 0.991 & 0.816 & 0.850 \\

\midrule

\textbf{EPnP-LM}  
& 0.437 & 0.207 & 0.163 & 0.510
& 0.605 & 0.483 & 0.864 & 0.916
& 0.070 & 0.613 & 0.081 & 0.565
& 33.053 & 118.480
& 1.000 & 0.851 & 0.539 & 0.853 & 0.811 \\

\textbf{GMLPnP}  
& 0.375 & 0.172 & 0.141 & 0.436
& 0.545 & 0.463 & 0.820 & 0.909
& 0.035 & 0.202 & 0.082 & 0.591
& 39.457 & 147.519
& 1.000 & 0.851 & 0.539 & 0.853 & 0.811 \\

\bottomrule
\end{tabular}
}

\vspace{1mm}
\begin{minipage}{0.98\textwidth}
\footnotesize
\textit{Note.} Values are averaged over 45 trials. The abbreviation \emph{w/o} denotes \emph{without}. 
\emph{EPnP-LM} and \emph{GMLPnP} are frame-wise PnP baselines with deterministic LED selection. 
Translation RMSEs are in meters, rotation RMSEs are in radians, and reprojection errors are in pixels. 
$e_t$ and $e_r$ denote aggregated translation and rotation RMSEs. 
$\bar{\Delta p}$, $Q_{95}(\Delta p)$, $\bar{\Delta \Theta}$, and $Q_{95}(\Delta \Theta)$ denote frame-to-frame smoothness metrics. 
$A_i$ is the ID agreement rate of LED $i$, and $\bar{A}$ is the mean ID agreement. 
Lower is better except for ID agreement.
\end{minipage}
\end{table*}

\subsection{Overall Results and Comparison}

Fig.~\ref{part1-1} presents a representative trial at the $(1.6,-0.4)$\,m test location, and Table~\ref{tab:overall_metrics} reports the averaged performance over all 45 trials. The figure illustrates temporal behavior in one typical sequence, while the table summarizes the overall quantitative comparisons across all experimental conditions.

\subsubsection{Ablation Analysis}

Fig.~\ref{part1-1} illustrates that PSwPnP closely follows the motion-capture ground truth in all six degrees of freedom. The translation estimates remain stable around the nominal relative position, while the yaw response follows the prescribed left and right $30^\circ$ rotations, approximately corresponding to $\pm 0.52$\,rad. In contrast, \emph{w/o EKF} introduces clear oscillations during turning. Around $15$\,s, when the leader is rotating and only three LEDs are reliable, the roll error of \emph{w/o EKF} approaches 2\,rad, indicating that partial LED observations become weakly constrained without the Lie-group rigid-body prediction. The variant \emph{w/o Assoc.} produces large translation and attitude deviations because incorrect LED correspondences corrupt the 2D--3D geometry, while \emph{w/o Exist.} becomes unstable when instantaneous visibility changes directly alter the reliable marker set.

Table~\ref{tab:overall_metrics} confirms these observations. PSwPnP achieves the lowest overall errors, with tranlation and rotation RMSEs of 0.032\,m and 0.032\,rad, respectively. Compared with \emph{w/o EKF}, the proposed method reduces the translation and rotation errors by 76.6\% and 94.5\%, respectively, while decreasing $Q_{95}(\Delta \Theta)$ from 1.932\,rad to 0.012\,rad. These results demonstrate that the Lie-group rigid-body predictor effectively preserves motion continuity and that suppresses frame-wise pose jitter under noisy and incomplete observations.

Replacing probabilistic association with deterministic matching (\emph{w/o Assoc.}) results in the largest performance degradation. Translation and rotation RMSEs increase to $0.499$\,m and $0.887$\,rad, respectively, accompanied by a substantial increase in reprojection error and a reduction in marker ID consistency. This demonstrates that reliable probabilistic association is essential for maintaining correct LED identities under underwater reflections, missed detections, and ambiguous marker observations.

Similarly, removing the existence probability (\emph{w/o Exist.}) significantly degrades both pose accuracy and identity consistency. Without temporal visibility modeling, the estimator reacts directly to instantaneous detection changes, causing frequent switches of the reliable marker set and unstable pose updates. The proposed existence-aware mechanism effectively suppresses these transient visibility fluctuations, resulting in substantially more stable pose estimation.

\subsubsection{Comparison With Frame-Wise PnP Methods}

Table~\ref{tab:overall_metrics} compares PSwPnP with two representative frame-wise PnP methods. Compared with EPnP-LM and GMLPnP, PSwPnP reduces the translation RMSE by 93.7\% and 92.7\%, respectively, while reducing the rotation RMSE by approximately 96.5\% for both methods.

These results indicate that robust underwater AUV-to-AUV relative pose estimation cannot be achieved by frame-wise PnP optimization alone. Although GMLPnP improves single-frame estimation by modeling measurement uncertainty, both baselines rely on deterministic four-marker correspondences and process each frame independently. In contrast, PSwPnP combines motion prediction, probabilistic marker association, existence-aware visibility management, and visibility-adaptive updates, resulting in substantially higher accuracy and temporal consistency.

\subsection{Marker Visibility Transition Study}

\begin{figure}[htbp]
    \centering
    \includegraphics[width=\linewidth, keepaspectratio]{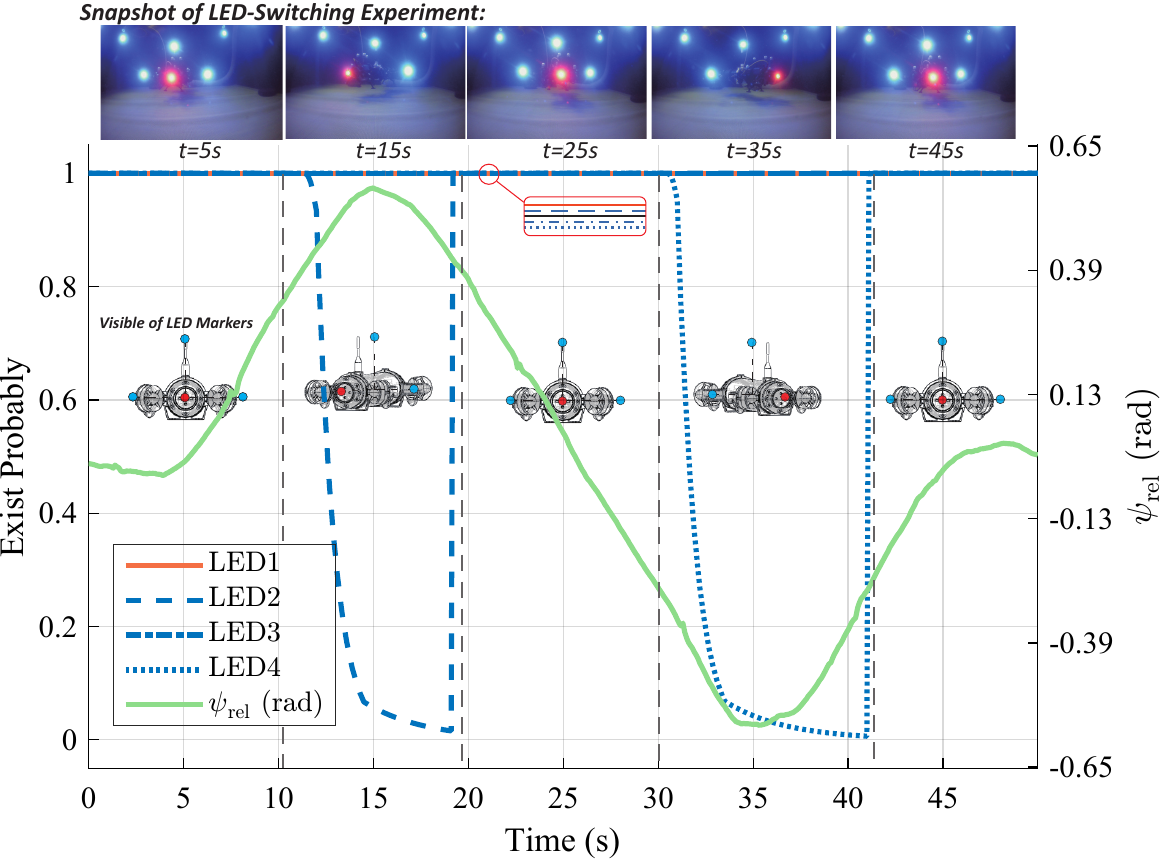}
    \caption{Representative $4$--$3$--$4$ marker visibility transition  at the $(1.4,0)$\,m test location during target yaw rotation.}
    \label{sw434}
\end{figure}

This section evaluates the $4$--$3$--$4$ marker visibility transition caused by self-occlusion. Fig.~\ref{sw434} gives one representative sequence, while Table~\ref{tab:switch_434_metrics} reports averaged results over all 45 trials.

A representative transition from the $(1.4,0)$\,m target location is illustrated in Fig.~\ref{sw434}. During the in-place yaw rotation, LED2 and LED4 alternately disappeared and reappeared, producing a typical $4$--$3$--$4$ visibility sequence.

\begin{table*}[htbp]
\centering
\caption{Performance comparison on the $4$--$3$--$4$ marker visibility transition intervals.}
\label{tab:switch_434_metrics}

\small
\setlength{\tabcolsep}{4.2pt}
\renewcommand{\arraystretch}{1.00}
\begin{tabular*}{0.88\textwidth}{@{\extracolsep{\fill}}lccccccccc@{}}
\toprule
\textbf{Method} 
& \multicolumn{3}{c}{\textbf{Transition Metrics}}
& \multicolumn{6}{c}{\textbf{Segment-wise RMSE}} \\
\cmidrule(lr){2-4}
\cmidrule(lr){5-10}
& $P_{4\rightarrow3}$ 
& $P_{3\rightarrow4}$ 
& $A_{\mathrm{mode}}$
& $e_t^{\mathrm{pre4}}$ 
& $e_r^{\mathrm{pre4}}$
& $e_t^{\mathrm{mid3}}$ 
& $e_r^{\mathrm{mid3}}$
& $e_t^{\mathrm{post4}}$ 
& $e_r^{\mathrm{post4}}$ \\
\midrule

\textbf{PSwPnP}      
& \textbf{0.899} & \textbf{0.956} & \textbf{0.941}
& \textbf{0.025} & \textbf{0.022}
& \textbf{0.036} & \textbf{0.042}
& \textbf{0.032} & \textbf{0.023} \\

\textbf{w/o EKF}    
& 0.899 & 0.951 & 0.940
& 0.033 & 0.113
& 0.257 & 1.134
& 0.051 & 0.155 \\

\textbf{w/o Assoc.}  
& 0.000 & 1.000 & 0.598
& 0.554 & 0.674
& 0.461 & 1.352
& 0.481 & 0.755 \\

\textbf{w/o Exist.}  
& 0.000 & 0.999 & 0.598
& 0.166 & 0.352
& 0.365 & 1.243
& 0.377 & 1.478 \\

\midrule

\textbf{EPnP-LM}  
& 0.000 & 0.999 & 0.598
& 0.536 & 0.701
& 0.444 & 1.387
& 0.508 & 0.650 \\

\textbf{GMLPnP}  
& 0.000 & 0.999 & 0.598
& 0.398 & 0.735
& 0.591 & 1.332
& 0.365 & 0.672 \\

\bottomrule
\end{tabular*}

\vspace{1mm}
\begin{minipage}{0.88\textwidth}
\footnotesize
\textit{Note.} Values are averaged over the annotated $4$--$3$--$4$ transition intervals from all 45 trials. 
\emph{EPnP-LM} and \emph{GMLPnP} are frame-wise PnP baselines with deterministic LED selection. 
$P_{4\rightarrow3}$, $P_{3\rightarrow4}$, and $A_{\mathrm{mode}}$ denote transition detection and mode accuracy metrics. 
The superscripts $\mathrm{pre4}$, $\mathrm{mid3}$, and $\mathrm{post4}$ denote the first 4-LED segment, the middle 3-LED segment, and the final 4-LED segment. 
Higher is better for transition metrics, while lower is better for RMSE.
\end{minipage}
\normalsize
\end{table*}

\subsubsection{Transition Performance}

Table~\ref{tab:switch_434_metrics} shows that PSwPnP maintains stable estimation during the visibility transition. It achieves $P_{4\rightarrow3}=0.899$, $P_{3\rightarrow4}=0.956$, and $A_{\mathrm{mode}}=0.941$. Its mid3 errors remain low at $e_t=0.036$\,m and $e_r=0.042$\,rad, demonstrating the effectiveness of the pixel-level tight-coupling update under partial observations.

Removing the Lie-group EKF (\emph{w/o EKF}) produces similar transition metrics but substantially larger mid3 errors ($e_t=0.257$\,m and $e_r=1.134$\,rad), indicating that correct visibility recognition alone is insufficient without motion prediction.

Removing probabilistic association or existence modeling causes transition failures ($P_{4\rightarrow3}=0$ and $A_{\mathrm{mode}}\approx0.60$), leading to rotation errors exceeding 1.2\,rad. The frame-wise EPnP-LM and GMLPnP baselines exhibit similar degradation because they neither preserve LED identities nor exploit partial observations.

\subsubsection{Effect of Existence Probability}

As illustrated in Fig.~\ref{sw434}, LED visibility changes gradually near transition boundaries because of self-occlusion and image noise. The proposed existence probability smooths frame-wise evidence into a stable visibility estimate, preventing transient detection failures from immediately changing the reliable LED set.
Consequently, PSwPnP switches reliably between pose-level and pixel-level updates, whereas hard gating (\emph{w/o Exist.}) frequently produces incorrect mode transitions and large pose errors.

\section{Closed-Loop Leader--Follower AUV Experiment}

To demonstrate real-time applicability, AMR-Pose was deployed in a closed-loop leader--follower experiment. The leader AUV carried the LED marker array, and the follower estimated the relative pose using a monocular camera and PSwPnP. The estimated pose was fed into a PID  tracking controller, whose objective was to maintain a relative distance of approximately 1.8\,m behind the leader along its body $x$-axis while aligning the follower yaw with the leader.

\begin{figure}[htbp]
    \centering
    \includegraphics[width=\linewidth, keepaspectratio]{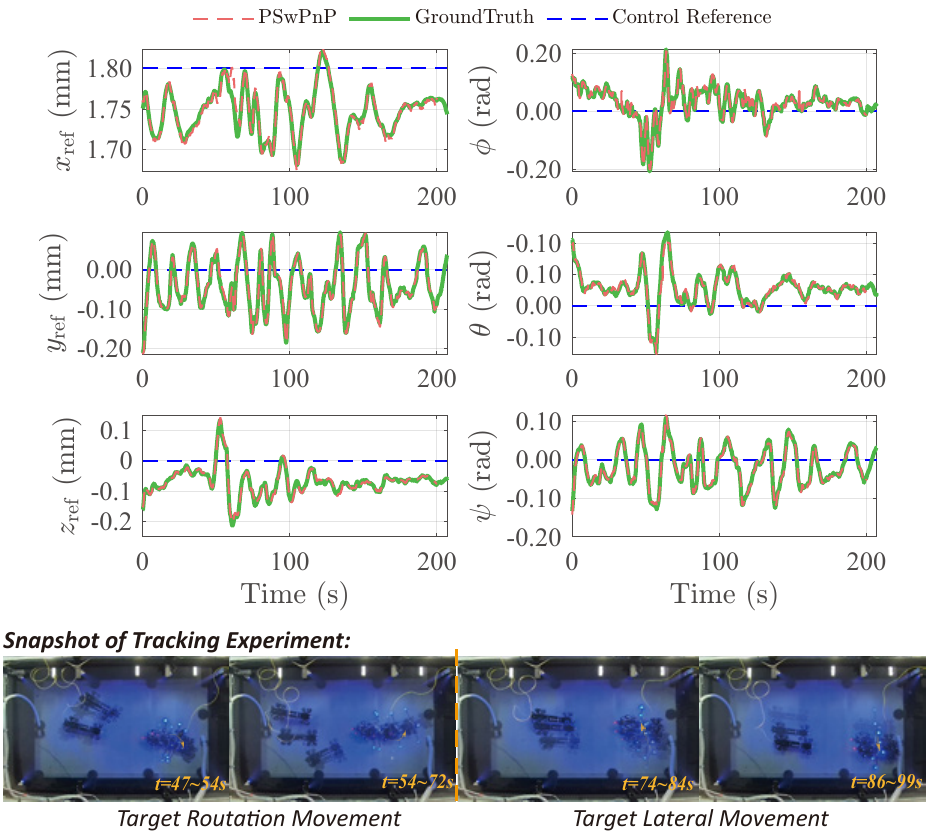}
    \caption{Closed-loop leader--follower experiment with AMR-Pose deployment.}
    \label{Tracking}
\end{figure}

Fig.~\ref{Tracking} shows that the estimated relative pose closely follows the motion-capture ground truth throughout leader maneuvers. The snapshots further illustrate stable closed-loop leader--follower tracking under representative translational and rotational motions, demonstrating the feasibility of integrating AMR-Pose into a real-time closed-loop perception-and-control pipeline for underwater leader--follower navigation.

\section{Conclusion}

This paper presented AMR-Pose, an active LED marker-based relative pose estimation framework for cooperative AUVs. The proposed system combines a compact red--blue LED marker array with the PSwPnP estimator to achieve robust 6-DoF relative pose tracking under marker ambiguity and partial visibility.
Extensive water-tank experiments demonstrated that AMR-Pose provides accurate, smooth, and robust relative pose estimation across diverse viewing conditions and $4$--$3$--$4$ marker visibility transitions. Closed-loop leader--follower experiments further demonstrated the feasibility of AMR-Pose for real-time underwater perception-and-control loop.

Future work will extend AMR-Pose to multi-AUV cooperative scenarios and improve its robustness in long-range, highly turbid, and unstructured underwater environments.

\bibliography{ref}

\begin{IEEEbiography}
    [{\includegraphics[width=1in,height=1.25in,clip,keepaspectratio]{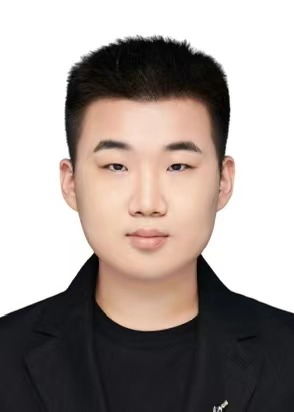}}]{Zeyu Sha} (Student Member, IEEE) 
    received the Bachelor's degree in Mechanical Design manufacture and Automation from Harbin University of Science and Technology, Harbin, China, in 2022 and the Master’s degree in Machinery in Department of Advanced Manufacturing and Robotics, College of Engineering, Peking University, Beijing, China, in 2025. He is currently working toward the Ph.D. degree in Department of Robotics Science and Engineering, School of Advanced Manufacturing and Robotics, Peking University, Beijing, China.
    
    His current research interests include mechatronic systems, robot design and manufacturing, and robotic system development.
\end{IEEEbiography}
\vspace{-15pt}

\begin{IEEEbiography}[{\includegraphics[width=1in,height=1.25in,clip,keepaspectratio]{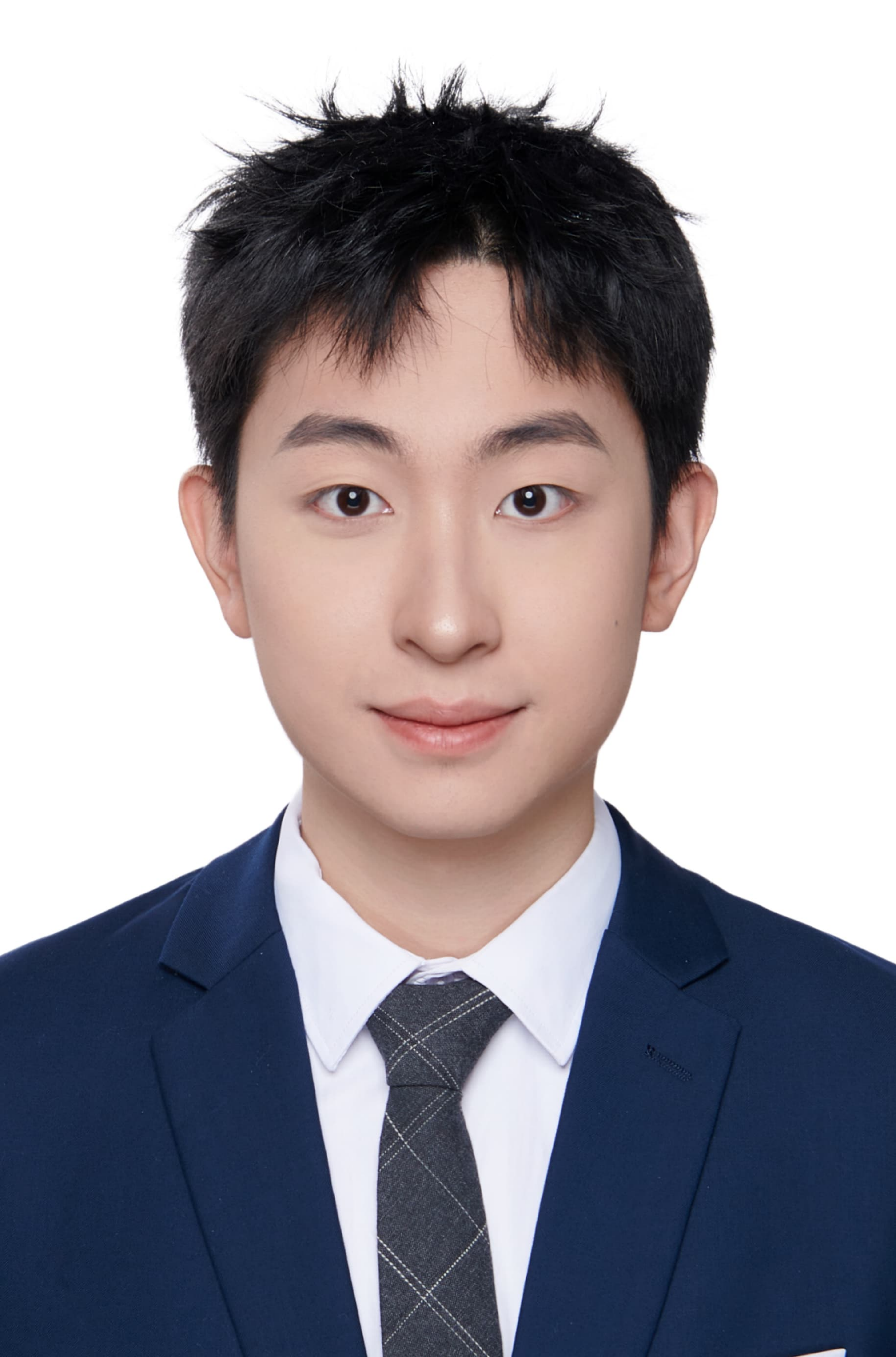}}]{Xiaorui Wang} (Student Member, IEEE) 
received the bachelor's degree in robotics engineering in 2025 from Peking University, Beijing, China, where he is currently working toward the Ph.D. degree in general mechanics and foundation of mechanics with the School of Advanced Manufacturing and Robotics, Peking University, Beijing, China. 

His research interests include underwater vehicles, dynamical modeling, and learning-based control.
\end{IEEEbiography}

\begin{IEEEbiography}
    [{\includegraphics[width=1in,height=1.25in,clip,keepaspectratio]{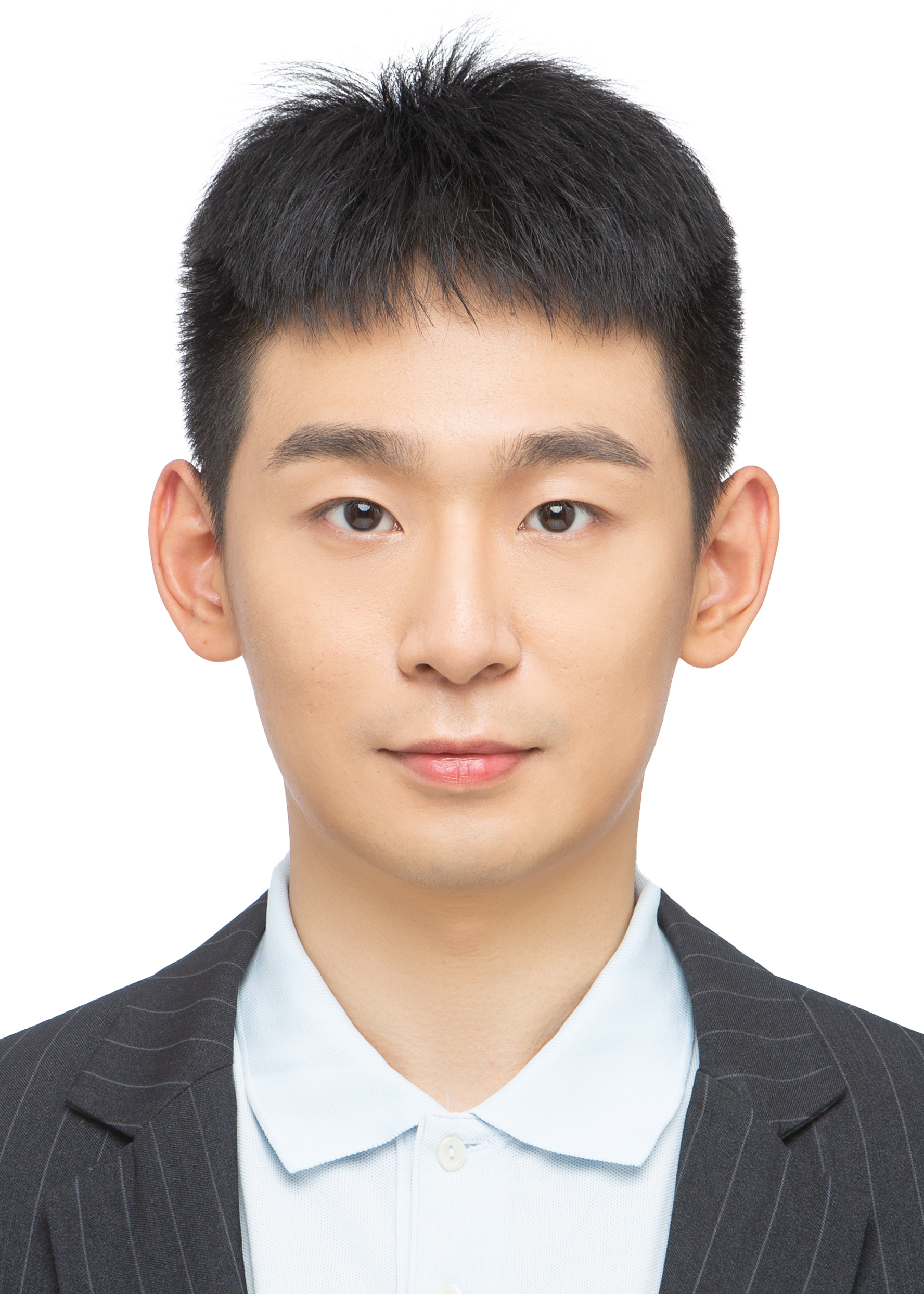}}]{Mingyang Yang} received the Bachelor’s degree in control science and engineering from Harbin Institute of Technology, Harbin, China, in 2017 and the Master’s degree in electrical and computer engineering from Cornell University, Ithaca, NY, USA, in 2018. He is currently working toward the Ph.D. degree in Department of Robotics Science and Engineering, School of Advanced Manufacturing and Robotics, Peking University, Beijing, China.
    
    His research interests include sensor fusion, underwater vehicles and human-robot interaction.
\end{IEEEbiography}
\vspace{-15pt}

\begin{IEEEbiography}
    [{\includegraphics[width=1in,height=1.25in,clip,keepaspectratio]{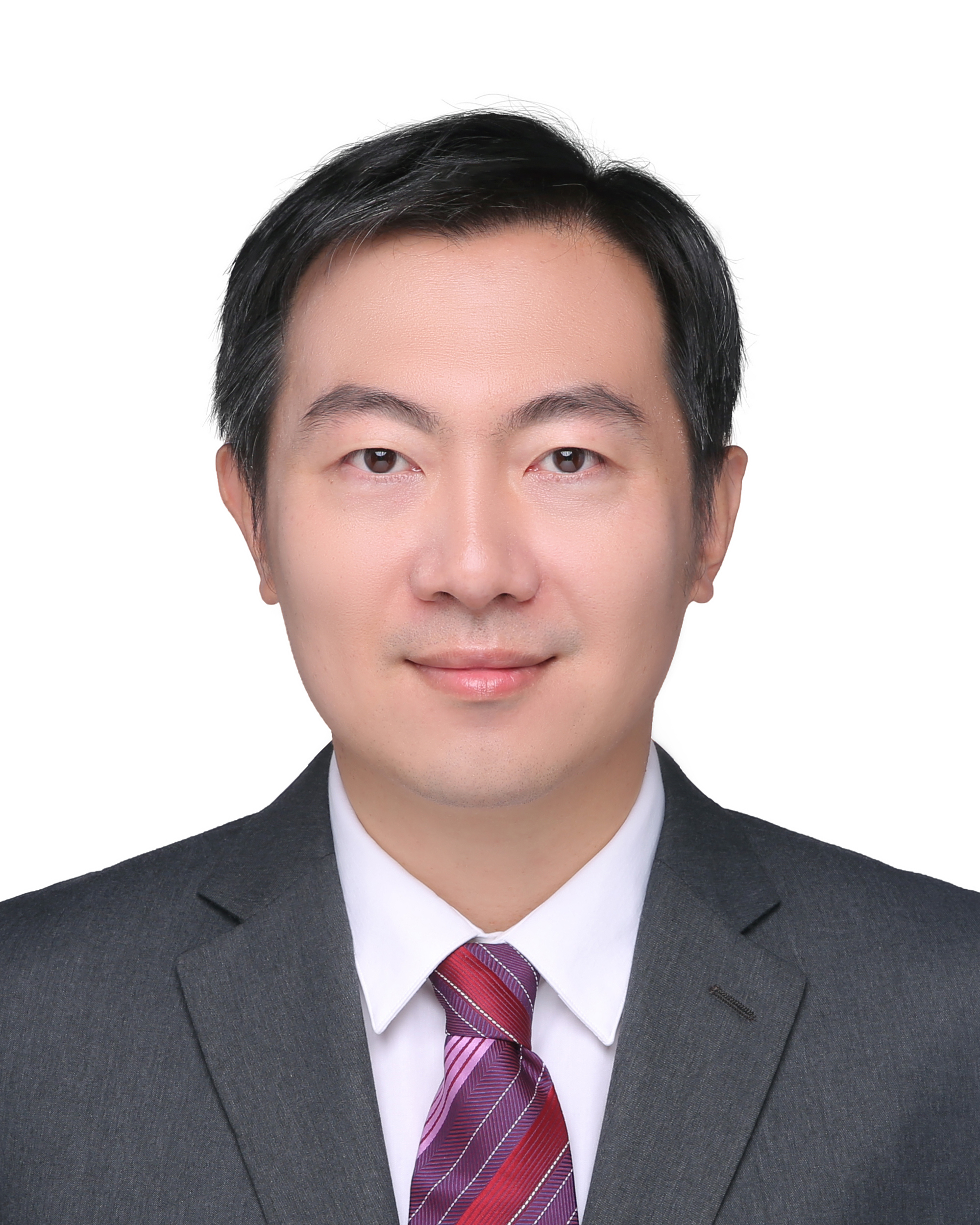}}]{Feitian Zhang} (S’12–M’14) received the Bachelor’s and Master’s degrees in automatic control from Harbin Institute of Technology, Harbin, China, in 2007 and 2009, respectively, and the Ph.D. degree in electrical and computer engineering from Michigan State University, East Lansing, MI, USA, in 2014. 
    
    From 2014 to 2016 and 2016 to 2021, he was a Postdoctoral Research Associate with the Department of Aerospace Engineering and Institute for Systems Research, University of Maryland, College Park, MD, USA, and an Assistant Professor of Electrical and Computer Engineering with George Mason University, Fairfax, VA, USA, respectively. 
    He is currently an Associate Professor of Robotics Engineering with Peking University, Beijing, China. His research interests include mechatronics systems, robotics and controls, aerial vehicles and underwater vehicles.
\end{IEEEbiography}

\end{document}